%% file: root.tex
\documentclass[letterpaper, 10 pt, conference]{ieeeconf}  

\IEEEoverridecommandlockouts                              
\usepackage{amsmath} 
\usepackage{amssymb}  
\usepackage{glossaries}
\usepackage{mathtools}
\usepackage{color}
\usepackage{cite}
\usepackage{pdfcomment}
\usepackage{booktabs}
\usepackage{adjustbox}
\usepackage{multicol}
\usepackage{pifont}
\usepackage{multirow}
\usepackage{subfig}
\usepackage{arydshln}

\newcommand{\cmark}{\ding{51}}%
\newcommand{\xmark}{\ding{55}}%

\newcommand{\etal}{\mbox{\emph{et al.~}}}
\newcommand{\figref}{Fig.~}
\newcommand{\tabref}{Tab.~}

\renewcommand{\baselinestretch}{0.95}
\DeclareMathOperator*{\softmaxA}{\textit{softmax}}  

\title{\LARGE \bf
Unknown Object Segmentation from Stereo Images
}

\author{Maximilian Durner*$^{1}$, Wout Boerdijk*$^{1}$, Martin Sundermeyer$^{1}$,\\ Werner Friedl$^{1}$, Zolt\'an Csaba M\'arton$^{1}$, Rudolph Triebel$^{1,2}$
\thanks{*Equal contribution}
\thanks{$^{1}$Institute of Robotics and Mechatronics, German Aerospace Center (DLR),
 	82234 Wessling, Germany
 	{\tt\small <first>.<second>@dlr.de}}%
\thanks{$^{2}$Department of Computer Science, Technical University of Munich (TUM),
 	85748 Garching, Germany}%
}

\begin{document}
\newglossaryentry{CNN}{name=CNN, description={Convolutional Neural Network},first={Convolutional Neural Network (CNN)}}
\newglossaryentry{UOIS}{name=UOIS, description={Unknown Object Instance Segmentation},first={\textit{Unknown Object Instance Segmentation} (UOIS)}}
\newglossaryentry{tfenc}{name=$TF_{Enc}$, description={Transformer Encoder},first={\textit{Transformer Encoder} ($TF_{Enc}$)}}
\newglossaryentry{tfdec}{name=$TF_{Dec}$, description={Transformer Decoder},first={\textit{Transformer Decoder} ($TF_{Dec}$)}}
\newglossaryentry{stios}{name=STIOS, description={Stereo Instances On Surfaces},first={\textit{Stereo Instances On Surfaces} (STIOS)}}
\newglossaryentry{instr}{name=INSTR, description={Instance Stereo Transformer},first={\textit{Instance Stereo Transformer } (INSTR)}}

\maketitle
\thispagestyle{empty}
\pagestyle{empty}

\begin{abstract}

Although instance-aware perception is a key prerequisite for many autonomous robotic applications, most of the methods only partially solve the problem by focusing solely on known object categories.
However, for robots interacting in dynamic and cluttered environments, this is not realistic and severely limits the range of potential applications.
Therefore, we propose a novel object instance segmentation approach that does not require any semantic or geometric information of the objects beforehand.
In contrast to existing works, we do not explicitly use depth data as input, but rely on the insight that slight viewpoint changes, which for example are provided by stereo image pairs, are often sufficient to determine object boundaries and thus to segment objects.
Focusing on the versatility of stereo sensors, we employ a transformer-based architecture that maps directly from the pair of input images to the object instances.  
This has the major advantage that instead of a noisy, and potentially incomplete depth map as an input, on which the segmentation is computed, we use the original image pair to infer the object instances and a dense depth map.
In experiments in several different application domains, we show that our \gls{instr} algorithm outperforms current state-of-the-art methods that are based on depth maps.
Training code and pretrained models will be made available.
\end{abstract}

\section{INTRODUCTION}
Robots interacting in real-world environments are often faced with a large variety of object instances.
Acquiring the information necessary for successful interaction with these objects is partly addressed by the field of object instance recognition, where large advances were made in terms of accuracy and robustness.
Nevertheless, the majority of existing methods require prior knowledge in terms of annotated data or 3D models for each considered object class or instance.

This work presents a novel stereo-based approach for \gls{UOIS} to address the mentioned issue in robotic vision.
Starting from the transformer encoder-decoder structure proposed in DETR~\cite{carion_end--end_2020} we modify the cross-attention mechanism in the decoder to directly predict instance segmentation masks without the intermediate detection step.
Our method, which is purely trained on synthetic images, is able to predict unknown objects on generic horizontal surfaces in an end-to-end manner without any post-processing due to the set-prediction attribute of the applied transformer structure.

\begin{figure}[!t]
	\centering 
	\includegraphics[width=1.\linewidth]{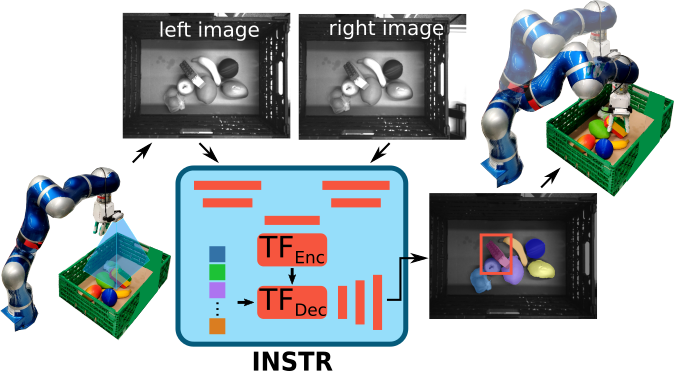} \caption{Given a stereo image pair INSTR segments unknown object instances on generic horizontal surfaces. The obtained pixel-wise object masks can be used for grasping, here exemplary with the clash hand~\cite{friedl2018clash}. The as auxiliary task predicted depth can further aide the robot in interacting with its environment.} 
	\label{fig:cover_fig}
\end{figure}%

As shown in~\cite{back_segmenting_2020, xiang_learning_2020} depth information is a crucial modality for robust \gls{UOIS}.
However, affordable depth sensors still cannot deal with untextured surfaces (stereo sensors), metallic, transparent or black materials and bright light (active sensors), which results in noisy and incomplete depth maps. 
Modeling imperfect depth data is difficult, sensor specific and thus often tackled by augmentations~\cite{back_segmenting_2020, xiang_learning_2020}. But are random augmentations on synthetic depth the best way for the network to decide when to rely on depth and when on RGB information at test time? Additionally, the fusion of RGB and corrupt depth data early in the network is non-trivial \cite{valada2017adapnet}. We hypothesize that a network that simultaneously learns disparity from stereo can build up a richer internal representation to determine where depth cues are useful and trustworthy and where it is better to follow textural information.

While depth data is beneficial in robotic applications, human-level segmentation of novel objects should be possible to learn from stereo image pairs alone, since we lack the capability of high-precision, active depth perception.
We calibrate our depth perception by reaching and walking \cite{Grzyb2013} well after we learn to recognize different objects, and are continuously correcting the errors we make during interaction with them \cite{Bingham2013}. \\
Concretely, we contribute the following:
\begin{itemize}
	\item We propose an end-to-end stereo-based approach that jointly learns disparity and unknown object instance segmentation from physically-based RGB stereo renderings. Experiments show that this is a promising approach to break the reliance on high quality depth data for robust unknown object segmentation. 
	\item We further introduce a sub-pixel sampling mechanism for our correlation layers, which enables a dynamic adaptation to other stereo sensor settings (e.g. changing baseline) the network was not trained on.
	\item We address the absence of an instance-centric stereo dataset and introduce \gls{stios}, a binocular dataset consisting of 192 scenes from two different stereo sensors on various surfaces with manually labelled, pixel-wise instance segmentation masks.
	\item We adapt the DETR\cite{carion_end--end_2020} transformer architecture for the task of instance segmentation.
	In concrete terms, by adapting the cross-attention mechanism as well as the segmentation loss, our transformer outputs 2D queries for direct upsampling.
		As result we obtain a post-processing-free class-agnostic instance segmentation pipeline running at 18 frames per second.
\end{itemize}

\section{RELATED WORK}
\textbf{Set-Prediction:}
Object-centric vision tasks such as (class-agnostic) detection or instance segmentation are usually tackled with CNN-based architectures. 
However, the arbitrary number of instances in a scene, their permutation invariance as well as the responsibility problem~\cite{zhang_deep_2019} are not directly solved by CNNs~\cite{carion_end--end_2020}. 
Instead, for these set-prediction problems, convolutional architectures usually generate a large number of proposals with hand-designed anchors that are subsequently refined.
Even anchor-free methods~\cite{duan2019centernet, tian2020fcos} then require expensive post-processing steps such as Non-Maximum Suppression~\cite{he_mask_2017}, watershed algorithm~\cite{bai_deep_2017}, hough-voting~\cite{xie_best_2020}, or clustering~\cite{de_brabandere_semantic_2017,neven_instance_2019, athar_stem-seg_2020} to filter the raw network outputs. 
Besides introducing additional hyperparameters, these post-processing methods take a large portion of the inference time when considering lightweight architectures that are crucial for robotic systems.

Direct, i.e post-processing free, set-prediction with CNNs requires adaptations, e.g. an autoregressive or recurrent structure to process scenes instance-by-instance~\cite{park_learning_2016, romera-paredes_recurrent_2016, ren_end--end_2017,  chen_learning_2018, salvador_recurrent_2019}.
Greff~\etal~\cite{greff_multi-object_2020} iteratively infer a set of latent representations, each representing an object, by variational inference.
In a similar manner multiple encoder-decoder steps are applied in~\cite{burgess_monet_2019, engelcke_genesis_2020} to refine object-centric representations.
Zhang~\etal~\cite{zhang_deep_2019} present a general set-prediction approach which optimizes the mean squared error between a latent representation of the input (image) and a set of feature vectors.
An iterative attention module is presented in~\cite{locatello_object-centric_2020} which groups task specific input features to a set of output vectors. While these experiments have indicated the potential of direct set-predictions, they haven't proven their applicability to complex real world applications, yet.

Recently, transformer networks~\cite{vaswani_attention_2017}, originally introduced for NLP tasks, gained a lot of attention in computer vision.
While the works \cite{dosovitskiy_image_2020,touvron_training_2021} apply transformers on sequences of image patches for classification, the DETR models~\cite{carion_end--end_2020, zhu_deformable_nodate} directly output a set of bounding box predictions in parallel by cross-attending object queries to the global image context.
An earlier work by Liang~\etal~\cite{liang_polytransform_2019} exploits self-attention across polygon vertices to improve the prediction of the offsets.
Based on the initial DETR model~\cite{carion_end--end_2020} several works have been published for instance segmentation\cite{prangemeier_attention-based_2020, wang_end--end_2020}.
These works reuse the original structure and predict the instance masks based on the bounding box features which resembles again an indirect approach.
The transformer network presented by Xie~\etal~\cite{xie_trans2seg_2021} addresses transparent object segmentation based on RGB. The object queries in their work encode class related features which is not applicable to the underlying task of unknown object segmentation.\\
\textbf{Unknown Object Instance Segmentation}: Early works in UOIS mainly build on low-level image features based on boundaries, connectivity or symmetry~\cite{pham_scenecut_2018, felzenszwalb_efficient_2004} to segment unknown instances. Richtsfeld~\etal~\cite{richtsfeld_segmentation_2012} proposed to estimate surface patches based on a mixture of planes and NURBS on which a graph-cut algorithm is applied.
However, such features are often insufficient to model what constitutes an object in more complex settings \cite{alexe_measuring_2012}. 
To let robots manipulate completely unknown instances we need to learn a concept of ``objectness'' defined as a geometrically and often semantically connected entity. One way to extract unknown objects from a scene is by predicting their independent motion masks \protect{\cite{xie_object_2019, dave_towards_2019}}. While most objects are naturally static, robotic manipulators can induce the necessary motion so that grasped objects can be segmented from arbitrary viewpoints \cite{boerdijk_self-supervised_2020}.
In~\cite{pinheiro_learning_2015} a class-agnostic segmentation mask together with an object likelihood score is predicted per RGB image patch of the MS COCO dataset~\cite{lin2014microsoft}.
However, color information from a few specific categories does not suffice to learn the ``objectness'' relevant in robotics contexts. 
For this purpose, more diverse training data can be generated synthetically by combining large 3D model databases like ShapeNet ~\cite{shapenet2015} with procedural data generation methods~\cite{denninger_blenderproc_2019}.
A straightforward approach is to train existing instance segmentation methods (e.g. Mask R-CNN~\cite{he_mask_2017}) with only a single foreground object category~\cite{danielczuk_segmenting_2019, back_segmenting_2020}. 
Xie~\etal~\cite{xie_best_2020} outperform this baseline by predicting 2D unit vectors pointing towards object centers from synthetic point clouds and then refining the predictions with another network using color information.
Similar to~\cite{de_brabandere_semantic_2017}, Xiang~\etal~\cite{xiang_learning_2020} cluster pixel-wise feature representation predicted by a single convolutional network jointly trained on synthetic RGB and depth data. 
Instead we jointly learn depth and unknown instance segmentation from photo-realistic stereo RGB renderings and directly predict instance masks with a transformer-based architecture.\\
\textbf{Stereo Segmentation:}
Recent networks are capable of sub-pixel accurate disparity estimation from stereo images \cite{cheng2020hierarchical}. Several works ~\cite{bleyer_object_2011, ladicky_joint_2010, ferrari_segstereo_2018, zhang_dispsegnet_2019, dovesi_real-time_2020} already show the mutual benefit of jointly learning disparity and semantic segmentation. In this work we investigate whether class-agnostic instance segmentation also benefits from jointly learning of disparity.

\section{METHOD}

\begin{figure*}[!t]
		\vspace{4mm}
	
	\begin{minipage}[t]{1.28\columnwidth}
		\vspace{0pt}
		\includegraphics[width=1.\columnwidth]{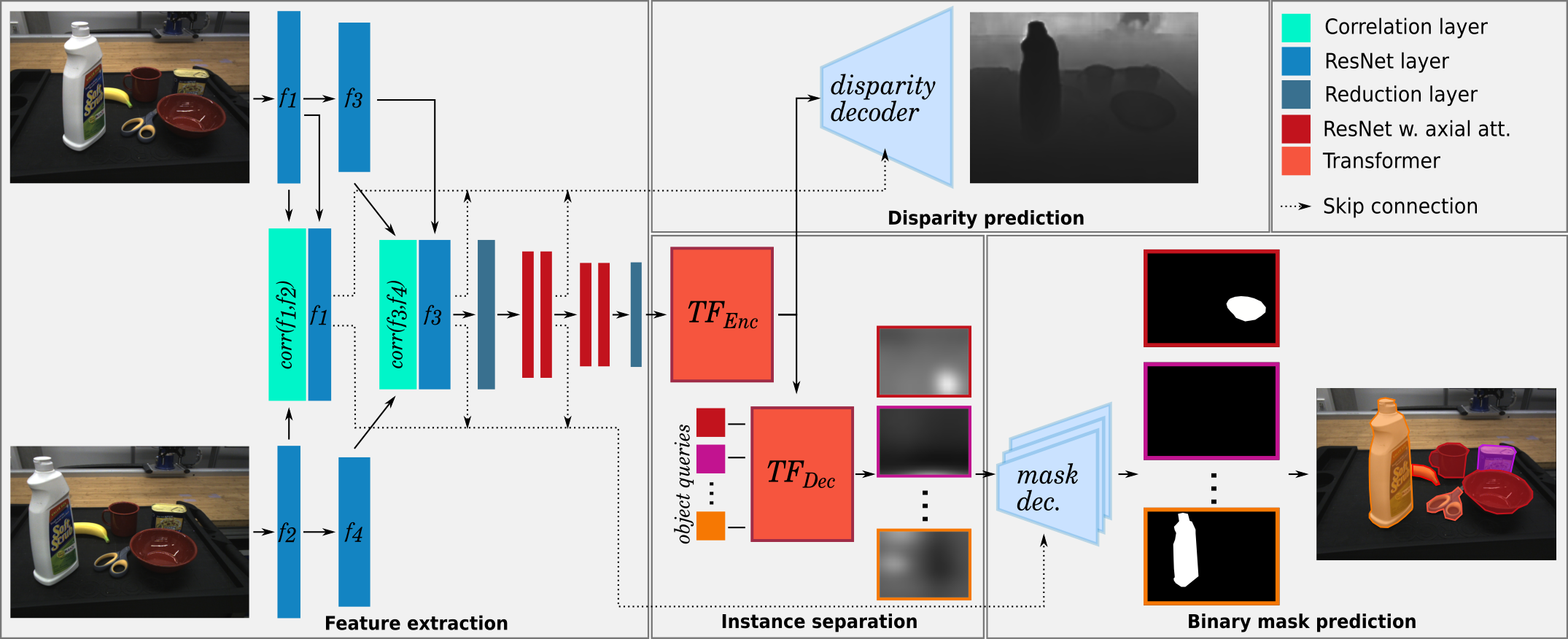}
		\captionof{figure}{Architecture of \gls{instr} (top); local horizontal correlation layer (right).}
		\label{fig:net_arch}
	\end{minipage}
	\hfill
	\begin{minipage}[t]{0.75\columnwidth}
		\vspace{0pt}
		\includegraphics[width=1.\columnwidth]{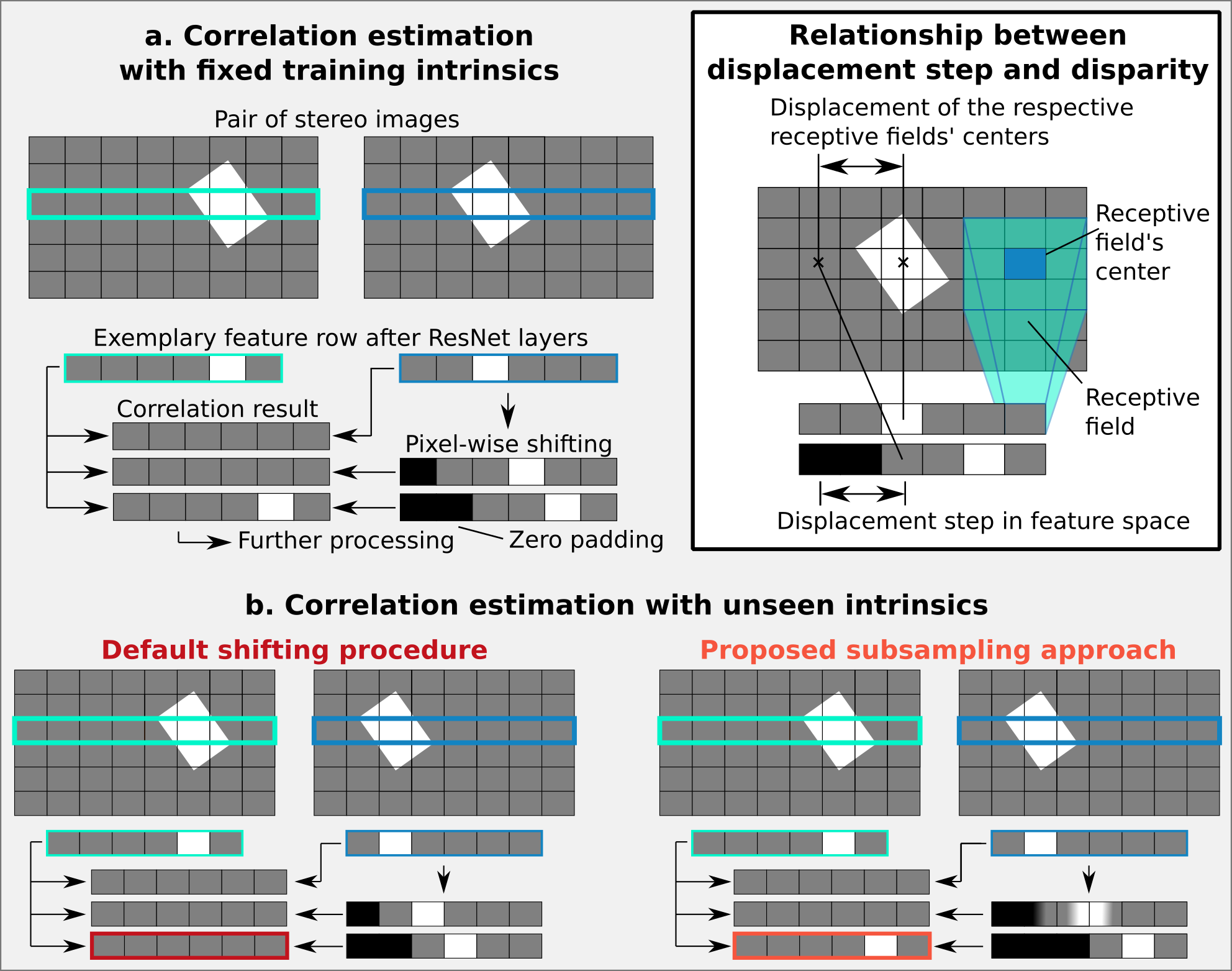}
	\end{minipage}
\end{figure*}

Our proposed method takes as input a pair of stereo-images and implicitly fuses the disparity and RGB cues to avoid the necessity of high-quality depth data.
As Fig.~\ref{fig:net_arch} shows, the inputs are forwarded through a feature extraction encoder to obtain a correlated feature representation of both images (Sec.~\ref{sec:feat_enc}).
Next, the features are processed by an instance-aware \gls{tfenc} followed by a \gls{tfdec} separating the instances (Sec.~\ref{sec:transformer}).
In contrast to the original~\gls{tfdec}, the one we apply outputs 2D feature maps which can directly be upsampled.
Also, another decoder is added to predict auxiliary disparities.

\subsection{Feature Extraction Encoder} \label{sec:feat_enc}
A stereo pair is forwarded through the first two layers of a ResNet-50 \cite{he2016deep} backbone with shared weights.
After each, a correlation layer (presented in~\cite{dosovitskiy2015flownet, ferrari_segstereo_2018}) restricted to a local horizontal region to capture stereoscopic information is applied. 
Mathematically, given the feature maps $\mathbf{f}_a, \mathbf{f}_b \in \mathbb{R}^{c\text{x}h\text{x}w}$ of a stereo pair with  $c, h, w$ correspond to number of channels, height, and width, we define the \emph{local horizontal correlation} at a specific spatial position $\mathbf{x}_a$ of $\mathbf{f}_a$ and $\mathbf{x}_b$ of $\mathbf{f}_b$ as

\begin{equation}
corr(\mathbf{x}_a, \mathbf{x}_b) = \sum_{\mathclap{i = 0}}^{d_{max}}\left\langle \mathbf{f}_a(\mathbf{x}_a),\mathbf{f}_b(\mathbf{x}_b+\begin{pmatrix}
i*s\\
0
\end{pmatrix})\right\rangle\;,
\end{equation}

with $d_{max}$ being the maximum shift of $\mathbf{f}_b$ in positive horizontal direction, and $s\!\!=\!\!1$ being the displacement size.
This displacement in a downsampled feature map can approximately be related to a certain disparity value given the centers of the respective receptive fields. 
Note that to have the same width as $\mathbf{f}_a$, $\mathbf{f}_b$ is zero padded on the left. 
The outcome is a correlation tensor $C\in \mathbb{R}^{c_c\text{x}h\text{x}w}$ where $c_c$ depicts the total number of displacement steps - a fixed value, since subsequent layers employ convolutional operations.
In consequence, the horizontal focal length $f_x$ and baseline $b_x$ are fixed during training. To dissolve this limitation and enable variable sensor intrinsics during inference, we allow continuous values for $s$ and obtain the corresponding $\mathbf{f}_b$ with bilinear grid sampling. 
Based on the relation with the receptive field, $d_{max}$ can be computed by:
\begin{equation}
d\textsubscript{max} \approx \frac{f_c *b_c}{z_{min}* output\_stride}\;,
\label{eq:dmax}
\end{equation}
where $z_{min}$ is the minimum camera-to-object distance and $output\_stride$ denotes the related downsampling ratio.
Given \eqref{eq:dmax} we then can generalize to arbitrary intrinsic parameters $f'_c$ and $b'_c$ by calculating the respective $d'_{max}$, and equally well compensate for a new $z'_{min}$.
The corresponding new step size $s'$ can be derived via the relation:
\begin{equation}
c_c = \frac{d\textsubscript{max}}{s} = \frac{d'\textsubscript{max}}{s'}\;.
\end{equation}

The intuition here is that given fixed intrinsics during training the network expects specific correlation results at specific channel positions of $c_c$.
Whenever $f_c$ and/or $b_c$ differ, the respective correlative features are at different positions and result in unexpected and confusing information to the network.
Changing $d_{max}$ and $s$ respectively counteracts to this behavior and maintains the features at their "correct" channel positions.
This adaptation of $d\textsubscript{max}$ and $s$ can be done without any retraining or fine-tuning since the correlation layer does not involve learnable parameters. 
Furthermore, the sub-pixel based sampling allows us to process arbitrary input sizes. 
Note that $d\textsubscript{max}$ is set such that the centers of the respective receptive fields cover more than the maximal (or any desired amount of) disparity, or a $z_{min}$.
Please see the right part of \figref\ref{fig:net_arch} for a visual explanation.

Since the correlation layer is computationally expensive the number of channels $c$ of $f_a$ and $f_b$ are reduced beforehand.
The correlation output of both layers is concatenated with the respective $f_a$ and passed as skip connection to the decoder.
For the second output the concatenated map is forwarded to subsequent encoder layers such that the transformer obtains correlation information.
We employ a reduction layer such that the channels fit the subsequent convolutional layer.

For segmentation tasks the stride of the following ResNet layers 3 and 4 is usually replaced with dilation to result in an $output\_stride$ of 8 instead of 32, which allows to extract denser feature responses \cite{chen2017rethinking}.
While this does not increase the number of learnable parameters in the encoder it results in an exponential increase of number of samples for the transformer (300$\rightarrow$4,800 for an image of 480x640), which would require more computational resources.
To capture meaningful features with $output\_stride\!=\!32$ we experiment with replacing ResNet layers 3 and 4 with their axial attention counterparts, and refer the reader to \cite{wang2020axial} for further details.
Finally, we channel-wise reduce the features (2048$\rightarrow$256) and pass them to the transformer.

\subsection{Transformer Encoder-Decoder} \label{sec:transformer}
The proposed approach builds up on the DETR architecture with an adapted \gls{tfdec}.
Based on a Transformer encoder-decoder structure, the main component is the attention mechanism, which is shortly described.
For further details we refer the reader to~\cite{carion_end--end_2020}.

Given a \textit{query sequence} $X_q \in \mathbb{R}^{t\text{x}N_q}$ and a \textit{key-value sequence} $X_{kv} \in \mathbb{R}^{t\text{x}N_{kv}}$, the embeddings query $Q$, key $K$ and value $V$ can be defined as linear projection: 
$Q\!=\!W_q (X_q+P_q);
K\!=\!W_k (X_{kv}+P_{k});
V\!\!=\!\!W_v X_{kv},$
where $W_k, W_q \in \mathbb{R}^{d_h\text{x}d}$ and $W_v \in \mathbb{R}^{d_{out}\text{x}d_{in}}$ are learned weights (in our case $d_{out}\!=\!d_{in}$; in the following both variables are denoted as $d$). 
The terms $P_q \in \mathbb{R}^{d\text{x}N_q}$ and $P_{k} \in \mathbb{R}^{d\text{x}N_{kv}}$, either learned or fixed, represent positional encodings to maintain spatial information, which is crucial for spatial structures or shapes.
In the next step an attention map $A$ is computed by the inner product between the query~$Q$ and key~$K$:
\begin{equation}
A = \softmaxA_{N_{kv}}(Q^T K)\;,
\end{equation}
where the softmax operation is applied along the $N_{kv}$ dimension.
The final output of one attention layer \texttt{att} is then computed by the attention-weighted sum over $V$.
Using the Einstein summation convention (\textit{einsum}) this can be expressed as:
\begin{equation}
\texttt{att}_{N_{q}d} = A_{N_{q}N_{kv}} V_{N_{kv}d}\;. \label{eq:att}
\end{equation}

For the attention mechanism applied in a $TF_{Enc}$ layer, called \textit{self-attention} all three embeddings are linear projections of the input map ($X_q\!=\!X_{kv}$).
On the other hand, the so-called \textit{cross-attention} mechanism $\texttt{c-att}$, applied in a \gls{tfdec} layer, attends the resulting $\texttt{att}$ as $X_{kv}$ with $Q$ consisting of a set of $N_q$ embeddings of length $t$.
The set-elements are learnt positional encoding, also called \emph{object queries}.
In the \gls{tfdec} every instance is represented by one object query based on the instance aware feature sequence generated by \gls{tfenc}.
Applying \eqref{eq:att} for cross-attention, as done originally, results in $N_q$ object queries each with dimension $d$.
While this works for detection, the resulting object queries cannot directly be used in a segmentation decoder as they only represent point-wise features.
To circumvent this, \cite{carion_end--end_2020} attend and concatenate these queries with the image tensor for the task of panoptic segmentation.
Similarly, \cite{xie_trans2seg_2021} upstream the attention weights of the last cross-attention.
Instead, we rewrite the \eqref{eq:att} to directly obtain an expanded attention-weighted feature map for each query:
\begin{equation}
	\texttt{c-att-exp}_{N_{q}N_{kv}d} = A_{N_{q}N_{kv}} V_{N_{kv}d}\;.
	\label{eq:cross-att}
\end{equation}
Summing over $N_{kv}$ results again in \texttt{att}~\eqref{eq:att}, used to forward information to the next \gls{tfdec} layer.
While for the auxiliary loss as well as for upsampling \texttt{c-att-exp} is considered.
Note, the number of computations of the expanded attention is the same as for the standard attention mechanism.
The final \gls{tfdec} outcome is a set of 2-D object queries 
$\hat{y} = \{\hat{y}_1,...,\hat{y}_{N_q}\}$ with $|\hat{y}|\!=\!N_q$, where $\hat{y_i}$ is either a binary mask representing an object or a zero mask.
As shown in \figref \ref{fig:net_arch} the object queries are then upsampled by independent \textit{mask decoders} with shared weights.
 
\subsection{Permutation-Invariant Instance Segmentation Loss}
Besides the permutation-invariance of \gls{tfdec}, the loss design is also crucial to enhance variety within the embedding set.
Given the final output-set $\hat{y}$ a bipartite matching against the ground truth masks $y=\{y_1,...,y_L\}$ has to be applied.
If $|\hat{y}| > |y|$ the difference is padded with zero ground truth masks.
In contrast to~\cite{carion_end--end_2020}, we directly apply the matching on the segmentation masks.
Given the dice loss~\cite{sudre_dice_2017} $\mathcal{L}_{dice} = 1 - \frac{2y\hat{y} + 1}{y + \hat{y} + 1}$ as cost function, the optimal assignment is computed by the Hungarian method~\cite{Kuhn1955thehungarian}.
Based on the outcome, the actual loss is calculated.
In this context a challenging task is to deal with the evaluation of zero masks, which is not considered by the original dice loss.
One solution is to treat the background as object by inverting the maps to $(1-\hat{y})$ and $(1-y)$.
However, the dice loss is favourable to big masks as a few incorrectly predicted pixels do not lead to a significant change of the loss value.
This prevents the network of predicting complete zero masks since almost perfect predictions already achieve decent loss values. 
To overcome this issue an exponential logarithmic dice loss~\cite{wong_3d_2018} to focus more on the least and most correct predictions is used.
For a query output $\hat{y_i}$ and its matched ground truth mask $\hat{y_i}$ it can be written as:
\begin{equation}
\mathcal{L}_{segm}^i = 
\begin{cases}
\mathcal{L}_{dice}(y_l^*,\hat{y_i}) & \text{if} \sum y_l^* \neq 0\\
-ln(\mathcal{L}_{dice}(y_l^*,\hat{y_i}))^{\gamma} & \text{if} \sum y_l^* = 0\\
\end{cases}  
\label{eq:dice}
\end{equation}

\subsection{Auxiliary Disparity Prediction}
In order to provide the network a guidance for an efficient exploitation of stereo cues, an auxiliary decoder is employed. 
As shown in \figref \ref{fig:net_arch} the feature map generated by \gls{tfenc} and intermediate features with correlation results 
as skip connections are used to predict the disparity map.
For this auxiliary task the \textit{Huber Loss} $L_{hub}$ \cite{huber1964robust} is employed and the total network loss can be written as 
\begin{equation}
\mathcal{L} = \alpha \mathcal{L}_{hub} + \beta \sum^{M_{dec}}_j \sum^{N_{q}}_i \mathcal{L}_{segm}^{ij}\;, \label{eq:total_loss}
\end{equation}
where $\alpha$ and $\beta$ are weighting factors, $M_{dec}$ is the number of layers in \gls{tfdec} and $\mathcal{L}_{segm}^{ij}$ indicates the segmentation loss of the $i$th query in the $j$th layer.

\section{stereo Instances on Surfaces Dataset}
There exist several datasets of objects on table-top surfaces, and many provide segmentation \cite{suchi2019easylabel, richtsfeld_segmentation_2012}, bounding box \cite{rennie2016dataset, sui2018never} or point cloud \cite{ecins2016cluttered} annotations.
Still, to the best of our knowledge none 
include both stereo images and pixelwise object instance annotations.
To address this issue we present \gls{stios} which consists of stereo images of objects on top of eight tabletop-like surfaces (white table, tool cabinet, conveyor belt, wooden plank, office carpet, wooden table, robot workbench, lab floor) which are situated in various environments.
We employ two stereo sensors, a rc\_visard \cite{rcvisard} and a Zed \cite{zedstereo}, which both capture depth information from stereo and, in the case of Zed, normal directions and point cloud data.
For each surface we manually select four camera positions to capture the scene at different distance ranges and elevation levels.
Every image depicts a configuration of four to six randomly sampled objects of a subset of the YCB Video dataset \cite{calli2017yale} \footnote{The following objects were used: \emph{003\_cracker\_box}, \emph{005\_tomato\_soup\_can}, \emph{006\_mustard\_bottle}, \emph{007\_tuna\_fish\_can}, \emph{008\_pudding\_box}, \emph{010\_potted\_meat\_can}, \emph{011\_banana}, \emph{019\_pitcher\_base}, \emph{021\_bleach\_cleanser}, \emph{024\_bowl}, \emph{025\_mug}, \emph{035\_power\_drill}, \emph{037\_scissors}, \emph{052\_extra\_large\_clamp}, \emph{061\_foam\_brick}.}, where object sampling considers the total number of occurrences per object which should be similar across all items.

Regarding object placement we follow the idea of previous work (e.g. \cite{suchi2019easylabel, richtsfeld_segmentation_2012}) and differentiate between \emph{simple} (no physical contact between objects) and \emph{difficult} (physically touching or stacked) scenes. 
Note that objects might appear occluded in the image plane irrespective of the setting.
For one camera pose we record three simple and three difficult configurations with both sensors, totaling in 6 images per camera pose and 24 images per surface area.
All configurations considered, the dataset consists of 192 stereo images for each sensor.
To improve the stereo matching for the rc\_visard we additionally project a pattern on the scene.
Every left image is manually annotated with corresponding ground truth object instance masks.
We believe that this dataset can serve as a reasonable baseline for stereo-aided object instance segmentation of table-top-like scenes.

\section{EXPERIMENTS}
\subsection{Simulating Objects on a Table}
Due to the lack of a suitable dataset,
synthetic training data is generated using BlenderProc~\cite{denninger_blenderproc_2019}.
In detail, we select table-top surfaces inside rooms of the SunCG dataset \cite{song2016ssc}, where five to twelve random instances of the ShapeNet dataset \cite{shapenet2015} are placed in a physical simulation\footnote{The respective config will be made publicly available.}.
For each table ten camera poses within the upper hemisphere of the table's center are sampled with varying distances.

\subsection{Implementation Details}
Our training data consists of 40,000 images (90/10 train/val split) with an input size of 480x640 pixels.
\tabref\ref{tab:corr_params} shows the settings for both correlation layers.
All experiments are conducted with $N_q=15$ (mainly due to memory limits), which in general should be larger than the total number of objects in a scene.
The exponential factor $\gamma$ of~\eqref{eq:dice} is set to 0.2, the loss~\eqref{eq:total_loss} is weighted with $\alpha\!\!=\!\!\beta\!\!=\!\!1$ and optimized by AdamW~\cite{loshchilov2019decoupled} with a weight decay of 1e-2. 
With a learning rate of 1e-4 for all trainable parameters and batch size of 2, 
training for 40 epochs roughly takes 4 days.
An inference forward pass on an Nvidia RTX 2080 takes around $55ms$, thus our algorithm can operate at roughly 18fps. 

\begin{table}[t]
	\vspace{3mm}
	\centering
	\caption[Correlation settings]{Parameter settings for both correlation layers.
		The parameter $d\textsubscript{max}$ corresponds to a minimum camera-to-object distance of $12cm$ for rc\_visard. \emph{r} depicts the downsampling ratio which reduces $c$ of $f_{\{a,b\}}$, and \emph{$\hat{D}$\textsubscript{max}} the approximate maximum disparity (see Sec.~\ref{sec:feat_enc}). All values denote pixels.}
	\vspace*{1mm}
	\resizebox{1.\columnwidth}{!}{
		\input{corr_params}}
	\label{tab:corr_params}
\end{table}

\subsection{Comparison to the State-of-the-art Methods}

\begin{figure*}[!t]
	\centering
	\subfloat[][Left RGB]{\includegraphics[width=0.125\textwidth]{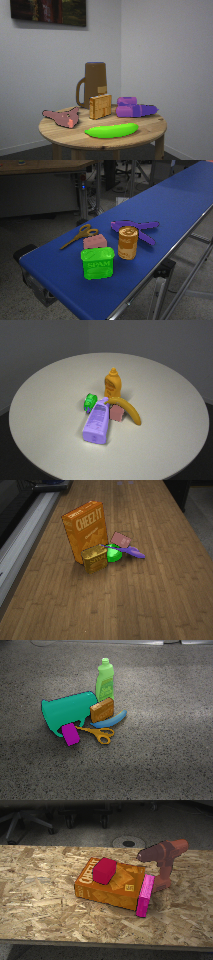}}
	\subfloat[][Depth w/o pat.]{\includegraphics[width=0.125\textwidth]{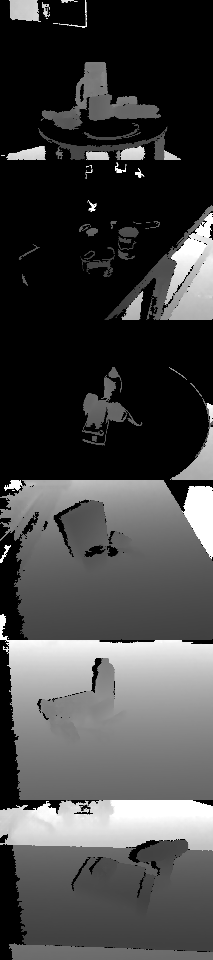}}
	\subfloat[][Depth w/ pat.]{\includegraphics[width=0.125\textwidth]{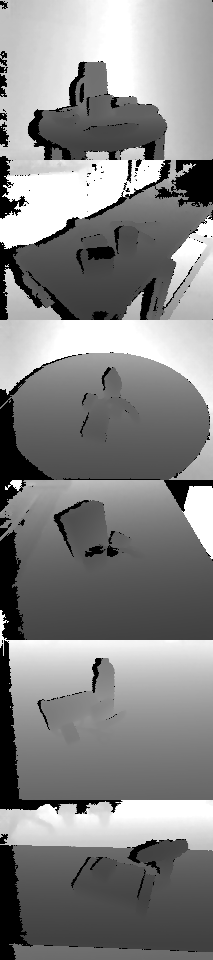}}
	\subfloat[][Xie \etal \cite{xie2020uois3d}]{\includegraphics[width=0.125\textwidth]{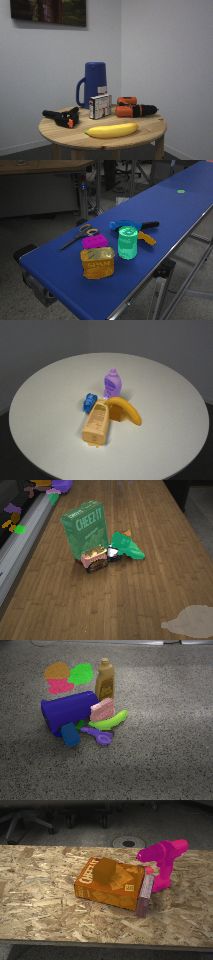}}	
	\subfloat[][Xiang \etal \cite{xiang_learning_2020}]{\includegraphics[width=0.125\textwidth]{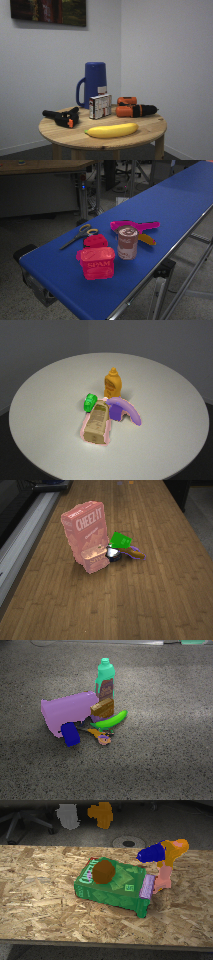}}
	\subfloat[][Mask-RCNN]{\includegraphics[width=0.125\textwidth]{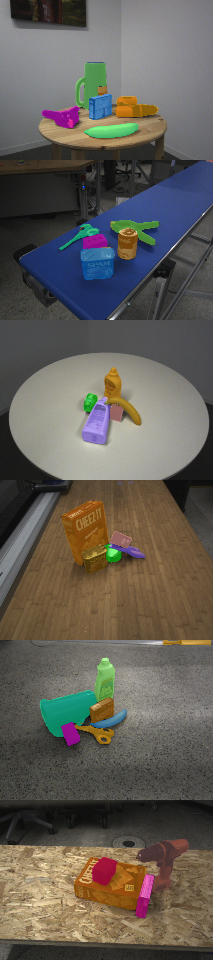}}
	\subfloat[][Ours]{\includegraphics[width=0.125\textwidth]{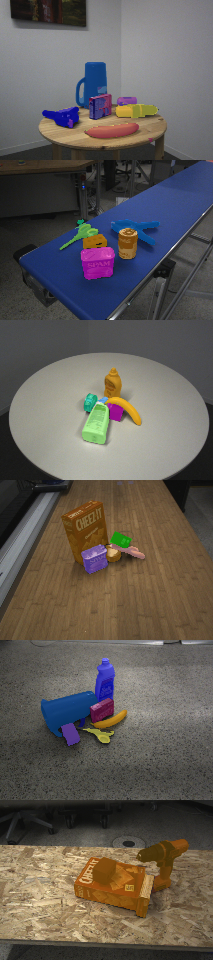}}
	\subfloat[][Our depth]{\includegraphics[width=0.125\textwidth]{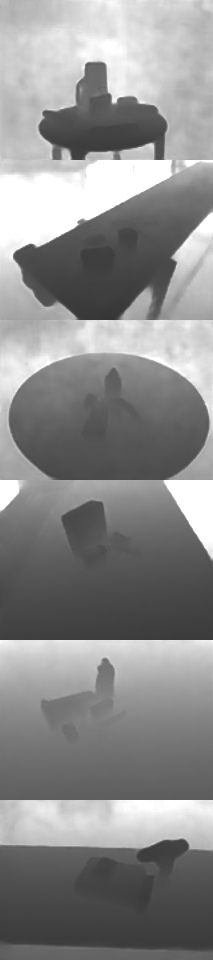}}
	\caption[Qualitative results]{Qualitative results on \gls{stios} (best viewed magnified and in color). Depth-based approaches trained on simulated data (d, e) struggle with real world stereo depth, especially if they contain fragmentary data (b, c). This potentially results in segmentation of background clutter (d: fourth / fifth row, e: last row), undetected objects (second row) or completely empty predictions (first row). \gls{instr} inherently utilizes cues from slight view point changes from a stereo pair and produces depth as auxiliary task. The last row denotes a particularly bad case where \gls{instr} fails to separate the objects. The predictions of d and e are based on depth with projected pattern. Colors are assigned randomly.}
	
	\label{fig:qualitative_results}
\end{figure*}

\tabref\ref{tab:quant_results} shows the comparison between our method and various recent baselines.
Evaluation is done by matching predicted objects with all ground truth instances of an image, and the set with less items is padded with empty masks.
As metrics we calculate the Intersection over Union (IoU) and F1 score on the matched pairs, and average across all images.
Xiang \etal~\cite{xiang_learning_2020} clusters object instances in deep feature space to circumvent set prediction, and \cite{xie2020uois3d} merges a foreground mask with object center vectors which is then refined by RGB data.
As an upper performance bound we train a MaskRCNN directly on photorealistically rendered 3D models of the YCB-V dataset used in the BOP Challenge~\cite{hodan2020bop}.
Exemplary qualitative results on \gls{stios} are depicted in \figref\ref{fig:qualitative_results}, and \figref\ref{fig:inthewild} shows the performance of \gls{instr} in the wild. 

\begin{table}[t]
	\vspace{3mm}
	\centering
	\caption[Quantitative results]{Mean IoU [\%] on object instance masks across all scenes of \gls{stios}. Values in \textbf{bold} denote the best results.}
	\vspace*{1mm}
	\resizebox{1.\columnwidth}{!}{
		\input{quant_results}}
	\label{tab:quant_results}
\end{table}

On STIOS the proposed method outperforms all baselines and achieves metrics en par with the Mask-RCNN that has already seen the objects.
While the projected pattern increases results for depth-based methods the obtained data is still insufficient for reliable predictions, possibly because of the amount of noise and incomplete data (total black areas of \figref\ref{fig:qualitative_results} b and c).

\subsection{Ablation Studies}
\label{sec:ablation}
\subsubsection{Architectural Ablation}
We continue by exploring the influence of various design choices to our network in \tabref\ref{tab:ablation}. 
Note that for all results in this section we only train up to 15 epochs and stop.
Both the axial attention blocks as well as employing an auxiliary loss on upsampled intermediate \gls{tfdec} outputs guides in learning meaningful representations; the latter being in coherence to previous work (e.g.~\cite{carion_end--end_2020}).
We furthermore compare our proposed query processing method \texttt{c-att-exp}~\eqref{eq:cross-att} with upsampling attention weight maps (\texttt{att},~\cite{xie_trans2seg_2021}) as well as concatenating these with backbone features (\texttt{c-att-cat-bb},~\cite{carion_end--end_2020}) and transformer encoder features (\texttt{c-att-cat-tfenc}).

\begin{table}[t]
	\centering
	\caption[Ablation study]{Architectural ablation study. All values denote mIoU [\%] and are computed after 15 epochs of training. Values in \textbf{bold} denote the best results.}
	\vspace*{1mm}
	\resizebox{1.\columnwidth}{!}{
		\input{ablation}}
	\label{tab:ablation}
\end{table}

\subsubsection{Varying Intrinsics During Inference}

\begin{figure}[!t]
	\centering 
	\includegraphics[width=1.\linewidth]{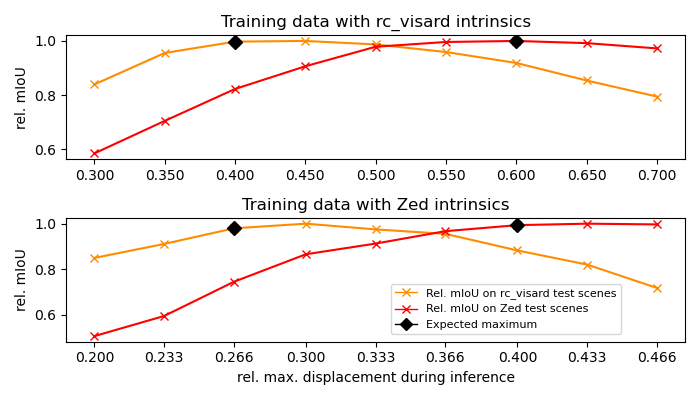} \caption{Relative mIoU on test scenes recorded with the rc\_visard (orange) and Zed (red) from training data with rc\_visard (top) and Zed intrinsics (bottom). Our correlation layer with subpixel sampling enables generalization to novel sensor intrinsics.}
	\label{fig:baseline_ablation}
	\vspace{-1.5em}
\end{figure}

The proposed correlation layer allows dynamic adjustment to different camera intrinsics (i.e., stereo baseline and the horizontal focal length).
To empirically validate this assumption two \gls{instr} models are trained with 5,000 synthetic samples with rc\_visard and Zed intrinsics, respectively.
In both trainings the maximum displacement is set to $d\textsubscript{max}= 0.4$.
\figref\ref{fig:baseline_ablation} depicts evaluation on both test sensors and verifies the ability to adapt to untrained sensor intrinsics. 

\subsubsection{Single-RGB \gls{instr}}
As shown in \tabref\ref{tab:input_modalities}, training \gls{instr} without the disparity loss yields higher accuracy than single RGB based predictions, which indicates that the local correlation itself presents a strong cue - albeit not as informative as with guidance in the form of a designated disparity loss.

\begin{table}[t]
	\centering
	\caption[Input modalities]{mIoU [\%] on synthetic validation and our real world test set given different input modalities.}
	\vspace*{1mm}
		\input{input_modalities}
	\label{tab:input_modalities}
\end{table}

\subsubsection{Depth Evaluation}

\begin{table}[t]
	\centering
	\caption[Depth evaluation]{L1 and RMS error [mm] evaluated on object regions of depth from rc\_visard with pattern.
	}
	\vspace*{1mm}
		\input{depth_evaluation}
	\label{tab:depth_evaluation}
\end{table}

As mentioned, \gls{instr} additionally predicts a pixel-wise disparity map.
Although this map only fulfils the task of auxiliary guidance, for completeness the L1 and the RMS error of the predicted disparity compared to the ground truth obtained from the rc\_visard with pattern are listed in \tabref\ref{tab:depth_evaluation}, and we also list the performance of AANet~\cite{xu2020aanet}, a dedicated stereo predictor, pretrained on Sceneflow~\cite{MIFDB16}.
Note that we only consider object ground truth regions and discard incomplete areas from calculation.
In addition, we experiment with utilizing the predicted depth of both~\cite{xu2020aanet} as well as \gls{instr} as input for the depth-only version of \cite{xiang_learning_2020}, and receive 33.42 and 17.97 \% mIoU on rc\_visard scenes of \gls{stios}.

\subsubsection{Single-RGB \gls{instr} on OCID Scenes}
For completion, we evaluate our single RGB based approach on the OCID~\cite{suchi2019easylabel} dataset in \tabref \ref{tab:ocid} aside the RGB-based comparison on our test dataset (\emph{Xiang} \etal \emph{RGB} in \tabref\ref{tab:quant_results} vs. \emph{RGB} in \tabref\ref{tab:input_modalities}).
We only consider scenes from ARID10 and YCB10 to have less objects present than our trained number of queries~($N_q~=~15$).
\begin{table}[t]
	\vspace{3mm}
	\centering
	\caption[Input modalities]{mIoU [\%] on two subsets (\emph{ARID10} and \emph{YCB10}) of the \emph{OCID}\cite{suchi2019easylabel} dataset.}
	\vspace*{1mm}
		\input{ocid}
	\label{tab:ocid}
\end{table}

\begin{figure}[!t]
	\centering 
	\includegraphics[width=1.\linewidth]{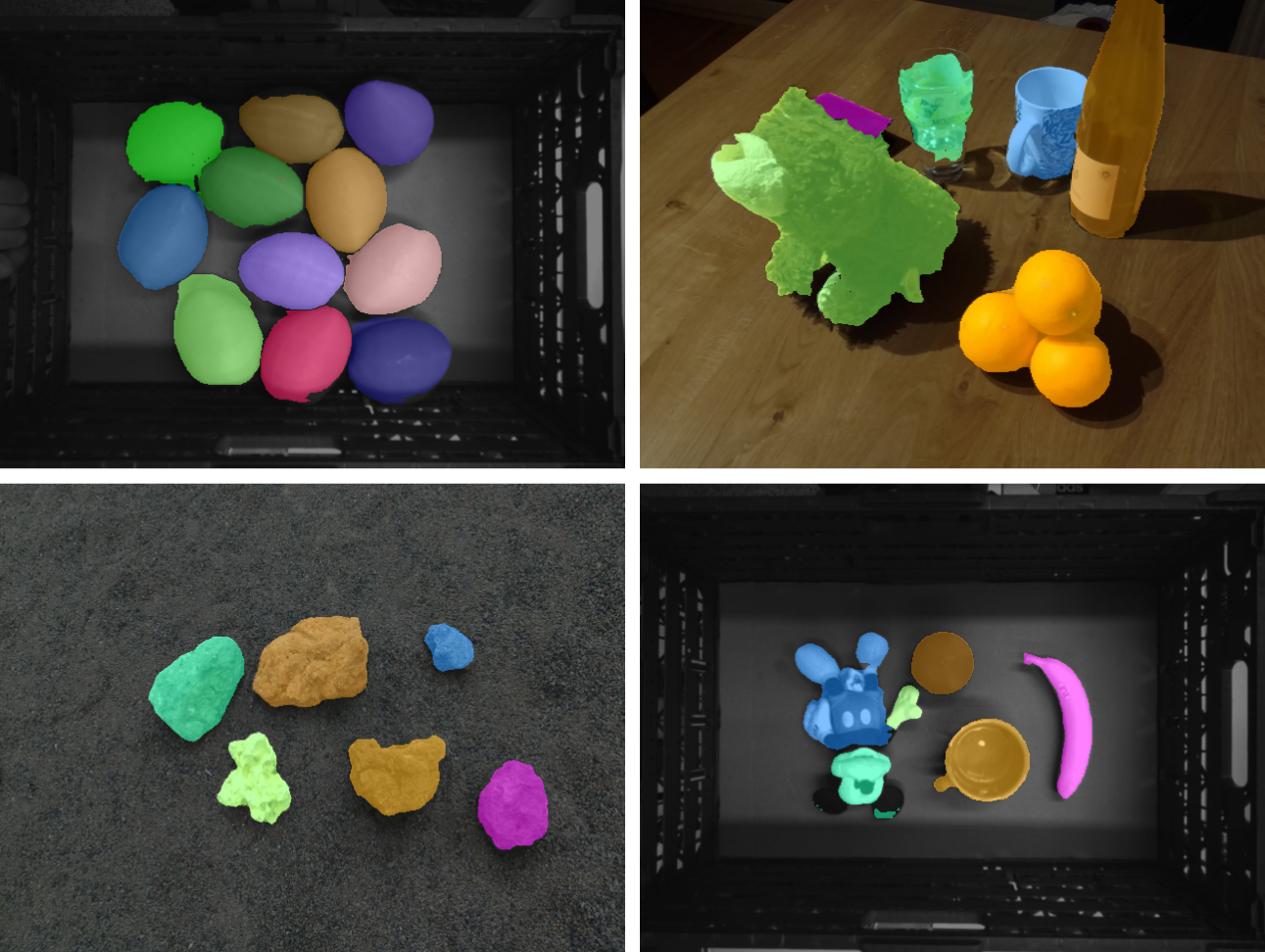} \caption{Further arbitrary objects on surfaces in the wild. While being trained on random ShapeNet instances on synthetic tables, \gls{instr} generalizes well to different domains: It segments mangos in a box (top left), transparent objects (top right), and robustly performs stone segmentation (bottom left). The Micky Mouse (bottom right) depicts a failure case where an object gets separated into multiple parts.}
	\label{fig:inthewild}
\end{figure}%

\section{CONCLUSIONS}
We have proposed \gls{instr}, a fast stereo-based instance segmentation approach (18fps) for unknown objects which addresses the issue of corrupted depth maps.
By applying local horizontal correlation the method is able to extract disparity-related as well as RGB-based features and learns a self-contained assessment of their significance.
Furthermore, the correlation mechanism applies sub-pixel sampling, which enables dynamic adaption to the underlying camera parameters.
Besides promising results on \gls{stios} we are able to grasp unseen objects as shown in \figref \ref{fig:cover_fig} (and further in the video), and can segment a variety of object shapes/textures in completely different domains (see \figref\ref{fig:inthewild}).
Exploiting binocular image pairs we hope to increase research interest towards robust stereo-aided robotic vision.

\







\bibliographystyle{IEEEtran} 
\small
\bibliography{bibfile}

\end{document}

%% file: corr_params.tex
\begin{tabular}{lcccccc}
\toprule
Layer & h/w & $d\textsubscript{max}$ & $r$ & $c_c$ &recept. field & $\hat{D}\textsubscript{max}$ \\

\midrule
Corr1 & 120/160 & 64 & 8 & 64 & 35x35 & 260 \\
Corr2 & 60/80 & 32 & 8 & 32 & 91x91 & 264\\

\bottomrule
\end{tabular}

%% file: quant_results.tex
\begin{tabular}{lcccccc}
\toprule
\multirow{2}{*}{Method} & \multicolumn{2}{c}{rc\_visard} & \multicolumn{2}{c}{rc\_visard + pattern} & \multicolumn{2}{c}{Zed}\\\cmidrule(lr){2-3} \cmidrule(lr){4-5} \cmidrule(lr){6-7} & mIoU & F1 & mIoU & F1 & mIoU & F1\\
\cmidrule(lr){1-1}\cmidrule(lr){2-5}\cmidrule(lr){6-7}

Xie \etal~\cite{xie2020uois3d} & 29.04 & 39.34 & 44.25 & 55.24 & 17.68 & 24.85\\
Xiang \etal~\cite{xiang_learning_2020} & 32.87 & 39.96 & 53.96 & 64.29 & 15.34 & 19.68\\
\hspace{3mm}RGB only & 53.44 & 66.78 & \textit{n/a} & \textit{n/a} & 49.43 & 62.39\\
\hspace{3mm}Depth only & 15.06 & 20.46 & 25.32 & 33.13 & 06.46 & 09.20\\

\gls{instr} & \textbf{74.92} & \textbf{84.49} & \textit{n/a} & \textit{n/a} & \textbf{74.31} & \textbf{84.07}\\
\hdashline[5pt/5pt]
\rule{0pt}{2.5ex}    Mask-RCNN (YCB-V) & 76.41 & 85.92 & \textit{n/a} & \textit{n/a} & 68.46 & 79.22\\
\bottomrule
\end{tabular}

%% file: ablation.tex
\begin{tabular}{cclccc}
\toprule
Ax. Bl. & Aux. loss & Query Proc. & val & rc\_visard & Zed \\
\midrule
\xmark & \xmark& c-att-exp &57.55&62.72 &57.57\\
\xmark & \cmark& c-att-exp &61.35& 67.48&63.60\\
\cmark & \xmark & c-att-exp &60.08 & 63.40 & 60.71\\
\midrule
\cmark & \cmark & c-att& 69.68 & 69.11 & 68.20 \\
\cmark  & \cmark &c-att-cat-bb& 69.30 & 68.31 & 64.15 \\ 
\cmark  & \cmark &c-att-cat-tfenc& 68.16 & 69.83 & 65.65 \\ 
\midrule
\textbf{\cmark} & \textbf{\cmark} & \textbf{c-att-exp}&\textbf{71.28} & \textbf{70.33} & \textbf{68.70} \\

\bottomrule
\end{tabular}

%% file: input_modalities.tex
\begin{tabular}{lccc}
\toprule
Input modality & Val & rc\_visard & Zed \\

\midrule
Single-RGB \gls{instr}& 69.30& 66.58& 63.52\\
\gls{instr} w/o disp. loss & 69.78 & 70.93 & 65.44\\
\gls{instr} & 77.46 & 74.92& 74.31\\

\bottomrule
\end{tabular}

%% file: depth_evaluation.tex
\begin{tabular}{lcc}
\toprule
Method & L1 error & RMS error\\
\midrule
Xu \etal~\cite{xu2020aanet} &53.80&59.08\\
\gls{instr} &54.33&58.55\\

\bottomrule
\end{tabular}	

%% file: ocid.tex
\begin{tabular}{lccc}
\toprule

Method & RGB & Depth & RGB+Depth\\
\midrule
Xiang \etal~\cite{xiang_learning_2020} &34.71&76.76&80.91\\
Single-RGB \gls{instr} &45.23&\textit{n/a}&\textit{n/a}\\
\bottomrule
\end{tabular}

%% file: root.bbl
\begin{thebibliography}{10}
\providecommand{\url}[1]{#1}
\csname url@rmstyle\endcsname
\providecommand{\newblock}{\relax}
\providecommand{\bibinfo}[2]{#2}
\providecommand\BIBentrySTDinterwordspacing{\spaceskip=0pt\relax}
\providecommand\BIBentryALTinterwordstretchfactor{4}
\providecommand\BIBentryALTinterwordspacing{\spaceskip=\fontdimen2\font plus
\BIBentryALTinterwordstretchfactor\fontdimen3\font minus
  \fontdimen4\font\relax}
\providecommand\BIBforeignlanguage[2]{{%
\expandafter\ifx\csname l@#1\endcsname\relax
\typeout{** WARNING: IEEEtran.bst: No hyphenation pattern has been}%
\typeout{** loaded for the language `#1'. Using the pattern for}%
\typeout{** the default language instead.}%
\else
\language=\csname l@#1\endcsname
\fi
#2}}

\bibitem{carion_end--end_2020}
N.~Carion, F.~Massa, G.~Synnaeve, N.~Usunier, A.~Kirillov, and S.~Zagoruyko,
  ``\BIBforeignlanguage{en}{End-to-{End} {Object} {Detection} with
  {Transformers}},'' \emph{\BIBforeignlanguage{en}{arXiv:2005.12872 [cs]}},
  2020.

\bibitem{friedl2018clash}
W.~Friedl, H.~H{\"o}ppner, F.~Schmidt, M.~A.~R. Garzon, and M.~Grebenstein,
  ``Clash: Compliant low cost antagonistic servo hands,'' in \emph{IEEE/RSJ
  International Conference on Intelligent Robots and Systems}, 2018.

\bibitem{back_segmenting_2020}
S.~Back, J.~Kim, R.~Kang, S.~Choi, and K.~Lee, ``Segmenting {Unseen}
  {Industrial} {Components} {In} {A} {Heavy} {Clutter} {Using} {RGB}-{D}
  {Fusion} {And} {Synthetic} {Data},'' in \emph{{IEEE} {International}
  {Conference} on {Image} {Processing} ({ICIP})}, 2020.

\bibitem{xiang_learning_2020}
Y.~Xiang, C.~Xie, A.~Mousavian, and D.~Fox, ``\BIBforeignlanguage{en}{Learning
  {RGB}-{D} {Feature} {Embeddings} for {Unseen} {Object} {Instance}
  {Segmentation}},'' \emph{\BIBforeignlanguage{en}{arXiv:2007.15157 [cs]}},
  2020.

\bibitem{valada2017adapnet}
A.~Valada, J.~Vertens, A.~Dhall, and W.~Burgard, ``Adapnet: Adaptive semantic
  segmentation in adverse environmental conditions,'' in \emph{IEEE
  International Conference on Robotics and Automation (ICRA)}, 2017.

\bibitem{Grzyb2013}
B.~J. {Grzyb}, L.~B. {Smith}, and A.~P. {del Pobil}, ``Reaching for the
  unreachable: Reorganization of reaching with walking,'' \emph{IEEE
  Transactions on Autonomous Mental Development}, 2013.

\bibitem{Bingham2013}
G.~P. Bingham and M.~A. Mon-Williams, ``The dynamics of sensorimotor
  calibration in reaching-to-grasp movements,'' \emph{Journal of
  Neurophysiology}.

\bibitem{zhang_deep_2019}
Y.~Zhang, J.~Hare, and A.~Prugel-Bennett, ``Deep {Set} {Prediction}
  {Networks},'' in \emph{Advances in {Neural} {Information} {Processing}
  {Systems}}, 2019.

\bibitem{duan2019centernet}
K.~Duan, S.~Bai, L.~Xie, H.~Qi, Q.~Huang, and Q.~Tian, ``Centernet: Keypoint
  triplets for object detection,'' in \emph{Proceedings of the IEEE/CVF
  International Conference on Computer Vision}, 2019.

\bibitem{tian2020fcos}
Z.~Tian, C.~Shen, H.~Chen, and T.~He, ``Fcos: A simple and strong anchor-free
  object detector,'' \emph{IEEE Transactions on Pattern Analysis and Machine
  Intelligence}, 2020.

\bibitem{he_mask_2017}
K.~He, G.~Gkioxari, P.~Doll{\'a}r, and R.~Girshick, ``Mask {R}-{CNN},'' in
  \emph{2017 {IEEE} {International} {Conference} on {Computer} {Vision}
  ({ICCV})}.

\bibitem{bai_deep_2017}
M.~Bai and R.~Urtasun, ``\BIBforeignlanguage{en}{Deep {Watershed} {Transform}
  for {Instance} {Segmentation}},''
  \emph{\BIBforeignlanguage{en}{arXiv:1611.08303 [cs]}}, 2017.

\bibitem{xie_best_2020}
C.~Xie, Y.~Xiang, A.~Mousavian, and D.~Fox, ``\BIBforeignlanguage{en}{The
  {Best} of {Both} {Modes}: {Separately} {Leveraging} {RGB} and {Depth} for
  {Unseen} {Object} {Instance} {Segmentation}},'' in
  \emph{\BIBforeignlanguage{en}{Conference on {Robotic} {Learning} ({CORL})}},
  2020.

\bibitem{de_brabandere_semantic_2017}
B.~De~Brabandere, D.~Neven, and L.~Van~Gool, ``\BIBforeignlanguage{en}{Semantic
  {Instance} {Segmentation} with a {Discriminative} {Loss} {Function}},'' in
  \emph{\BIBforeignlanguage{en}{{arXiv}:1708.02551 [cs]}}, 2017.

\bibitem{neven_instance_2019}
D.~Neven, B.~D. Brabandere, M.~Proesmans, and L.~Van~Gool,
  ``\BIBforeignlanguage{en}{Instance {Segmentation} by {Jointly} {Optimizing}
  {Spatial} {Embeddings} and {Clustering} {Bandwidth}},'' in
  \emph{\BIBforeignlanguage{en}{2019 {IEEE}/{CVF} {Conference} on {Computer}
  {Vision} and {Pattern} {Recognition} ({CVPR})}}.

\bibitem{athar_stem-seg_2020}
A.~Athar, S.~Mahadevan, A.~O{\v s}ep, L.~Leal-Taix{\'e}, and B.~Leibe,
  ``\BIBforeignlanguage{en}{{STEm}-{Seg}: {Spatio}-temporal {Embeddings} for
  {Instance} {Segmentation} in {Videos}},'' in
  \emph{\BIBforeignlanguage{en}{Computer {Vision} {\textendash} {ECCV} 2020}},
  Aug. 2020.

\bibitem{park_learning_2016}
E.~Park and A.~C. Berg, ``\BIBforeignlanguage{en}{Learning to decompose for
  object detection and instance segmentation},''
  \emph{\BIBforeignlanguage{en}{arXiv:1511.06449 [cs]}}.

\bibitem{romera-paredes_recurrent_2016}
B.~Romera-Paredes and P.~H.~S. Torr, ``\BIBforeignlanguage{en}{Recurrent
  {Instance} {Segmentation}},'' in \emph{\BIBforeignlanguage{en}{Computer
  {Vision} {\textendash} {ECCV} 2016}}.

\bibitem{ren_end--end_2017}
M.~Ren and R.~S. Zemel, ``\BIBforeignlanguage{en}{End-to-{End} {Instance}
  {Segmentation} with {Recurrent} {Attention}},'' in
  \emph{\BIBforeignlanguage{en}{{IEEE} {Conference} on {Computer} {Vision} and
  {Pattern} {Recognition} ({CVPR})}}, 2017.

\bibitem{chen_learning_2018}
T.~Chen, L.~Lin, X.~Wu, N.~Xiao, and X.~Luo, ``\BIBforeignlanguage{en}{Learning
  to {Segment} {Object} {Candidates} via {Recursive} {Neural} {Networks}},''
  \emph{\BIBforeignlanguage{en}{IEEE Transactions on Image Processing}}, 2018.

\bibitem{salvador_recurrent_2019}
A.~Salvador, M.~Bellver, V.~Campos, M.~Baradad, F.~Marques, J.~Torres, and
  X.~Giro-i Nieto, ``\BIBforeignlanguage{en}{Recurrent {Neural} {Networks} for
  {Semantic} {Instance} {Segmentation}},''
  \emph{\BIBforeignlanguage{en}{arXiv:1712.00617 [cs]}}, 2019.

\bibitem{greff_multi-object_2020}
K.~Greff, R.~L. Kaufman, R.~Kabra, N.~Watters, C.~Burgess, D.~Zoran,
  L.~Matthey, M.~Botvinick, and A.~Lerchner,
  ``\BIBforeignlanguage{en}{Multi-{Object} {Representation} {Learning} with
  {Iterative} {Variational} {Inference}},''
  \emph{\BIBforeignlanguage{en}{arXiv:1903.00450 [cs, stat]}}, 2020.

\bibitem{burgess_monet_2019}
C.~P. Burgess, L.~Matthey, N.~Watters, R.~Kabra, I.~Higgins, M.~Botvinick, and
  A.~Lerchner, ``\BIBforeignlanguage{en}{{MONet}: {Unsupervised} {Scene}
  {Decomposition} and {Representation}},''
  \emph{\BIBforeignlanguage{en}{arXiv:1901.11390 [cs, stat]}}, 2019.

\bibitem{engelcke_genesis_2020}
M.~Engelcke, A.~R. Kosiorek, O.~P. Jones, and I.~Posner,
  ``\BIBforeignlanguage{en}{{GENESIS}: {Generative} {Scene} {Inference} and
  {Sampling} with {Object}-{Centric} {Latent} {Representations}},''
  \emph{\BIBforeignlanguage{en}{arXiv:1907.13052 [cs, stat]}}, 2020.

\bibitem{locatello_object-centric_2020}
F.~Locatello, D.~Weissenborn, T.~Unterthiner, A.~Mahendran, G.~Heigold,
  J.~Uszkoreit, A.~Dosovitskiy, and T.~Kipf,
  ``\BIBforeignlanguage{en}{Object-{Centric} {Learning} with {Slot}
  {Attention}},'' \emph{\BIBforeignlanguage{en}{arXiv:2006.15055 [cs, stat]}},
  2020.

\bibitem{vaswani_attention_2017}
A.~Vaswani, N.~Shazeer, N.~Parmar, J.~Uszkoreit, L.~Jones, A.~N. Gomez, {\L
  }.~Kaiser, and I.~Polosukhin, ``Attention is all you need,'' in
  \emph{Proceedings of the 31st {International} {Conference} on {Neural}
  {Information} {Processing} {Systems}}, 2017.

\bibitem{dosovitskiy_image_2020}
A.~Dosovitskiy, L.~Beyer, A.~Kolesnikov, D.~Weissenborn, X.~Zhai,
  T.~Unterthiner, M.~Dehghani, M.~Minderer, G.~Heigold, S.~Gelly, J.~Uszkoreit,
  and N.~Houlsby, ``\BIBforeignlanguage{en}{An {Image} is {Worth} 16x16
  {Words}: {Transformers} for {Image} {Recognition} at {Scale}},''
  \emph{\BIBforeignlanguage{en}{arXiv:2010.11929 [cs]}}, 2020.

\bibitem{touvron_training_2021}
H.~Touvron, M.~Cord, M.~Douze, F.~Massa, A.~Sablayrolles, and H.~J{\'e}gou,
  ``\BIBforeignlanguage{en}{Training data-efficient image transformers \&
  distillation through attention},''
  \emph{\BIBforeignlanguage{en}{arXiv:2012.12877 [cs]}}, 2021.

\bibitem{zhu_deformable_nodate}
X.~Zhu, H.~Hu, S.~Lin, and J.~Dai, ``\BIBforeignlanguage{en}{Deformable
  {ConvNets} {V2}: {More} {Deformable}, {Better} {Results}},'' in
  \emph{\BIBforeignlanguage{en}{Proceedings of the IEEE/CVF Conference on
  Computer Vision and Pattern Recognition}}, 2019.

\bibitem{liang_polytransform_2019}
J.~Liang, N.~Homayounfar, W.-C. Ma, Y.~Xiong, R.~Hu, and R.~Urtasun,
  ``\BIBforeignlanguage{en}{{PolyTransform}: {Deep} {Polygon} {Transformer} for
  {Instance} {Segmentation}},'' \emph{\BIBforeignlanguage{en}{arXiv:1912.02801
  [cs]}}, 2019.

\bibitem{prangemeier_attention-based_2020}
T.~Prangemeier, C.~Reich, and H.~Koeppl, ``Attention-{Based} {Transformers} for
  {Instance} {Segmentation} of {Cells} in {Microstructures},'' in \emph{{IEEE}
  {International} {Conference} on {Bioinformatics} and {Biomedicine}}, 2020.

\bibitem{wang_end--end_2020}
Y.~Wang, Z.~Xu, X.~Wang, C.~Shen, B.~Cheng, H.~Shen, and H.~Xia,
  ``\BIBforeignlanguage{en}{End-to-{End} {Video} {Instance} {Segmentation} with
  {Transformers}},'' \emph{\BIBforeignlanguage{en}{arXiv:2011.14503 [cs]}},
  2020.

\bibitem{xie_trans2seg_2021}
E.~Xie, W.~Wang, W.~Wang, P.~Sun, H.~Xu, D.~Liang, and P.~Luo,
  ``\BIBforeignlanguage{en}{{Trans2Seg}: {Transparent} {Object} {Segmentation}
  with {Transformer}},'' \emph{\BIBforeignlanguage{en}{arXiv:2101.08461 [cs]}},
  2021.

\bibitem{pham_scenecut_2018}
T.~Pham, T.-T. Do, N.~S{\"u}nderhauf, and I.~Reid,
  ``\BIBforeignlanguage{en}{{SceneCut}: {Joint} {Geometric} and {Object}
  {Segmentation} for {Indoor} {Scenes}},''
  \emph{\BIBforeignlanguage{en}{arXiv:1709.07158 [cs]}}, 2018.

\bibitem{felzenszwalb_efficient_2004}
P.~F. Felzenszwalb and D.~P. Huttenlocher, ``\BIBforeignlanguage{en}{Efficient
  {Graph}-{Based} {Image} {Segmentation}},''
  \emph{\BIBforeignlanguage{en}{International Journal of Computer Vision}},
  2004.

\bibitem{richtsfeld_segmentation_2012}
A.~Richtsfeld, T.~M{\"o}rwald, J.~Prankl, M.~Zillich, and M.~Vincze,
  ``Segmentation of unknown objects in indoor environments,'' in
  \emph{{IEEE}/{RSJ} {International} {Conference} on {Intelligent} {Robots} and
  {Systems}}, 2012.

\bibitem{alexe_measuring_2012}
B.~Alexe, T.~Deselaers, and V.~Ferrari, ``\BIBforeignlanguage{en}{Measuring the
  {Objectness} of {Image} {Windows}},'' \emph{\BIBforeignlanguage{en}{IEEE
  Transactions on Pattern Analysis and Machine Intelligence}}, 2012.

\bibitem{xie_object_2019}
C.~Xie, Y.~Xiang, Z.~Harchaoui, and D.~Fox, ``\BIBforeignlanguage{en}{Object
  {Discovery} in {Videos} as {Foreground} {Motion} {Clustering}},'' in
  \emph{\BIBforeignlanguage{en}{{IEEE}/{CVF} {Conference} on {Computer}
  {Vision} and {Pattern} {Recognition} ({CVPR})}}, 2019.

\bibitem{dave_towards_2019}
A.~Dave, P.~Tokmakov, and D.~Ramanan, ``\BIBforeignlanguage{en}{Towards
  {Segmenting} {Anything} {That} {Moves}},'' in
  \emph{\BIBforeignlanguage{en}{{IEEE}/{CVF} {International} {Conference} on
  {Computer} {Vision} {Workshop} ({ICCVW})}}, 2019.

\bibitem{boerdijk_self-supervised_2020}
W.~Boerdijk, M.~Sundermeyer, M.~Durner, and R.~Triebel,
  ``\BIBforeignlanguage{en}{Self-{Supervised} {Object}-in-{Gripper}
  {Segmentation} from {Robotic} {Motions}},'' in
  \emph{\BIBforeignlanguage{en}{Conference on {Robotic} {Learning} ({CORL})}},
  2020.

\bibitem{pinheiro_learning_2015}
P.~O. Pinheiro, R.~Collobert, and P.~Doll{\'a}r, ``Learning to segment object
  candidates,'' in \emph{Proceedings of the 28th {International} {Conference}
  on {Neural} {Information} {Processing} {Systems}}, 2015.

\bibitem{lin2014microsoft}
T.-Y. Lin, M.~Maire, S.~Belongie, J.~Hays, P.~Perona, D.~Ramanan,
  P.~Doll{\'a}r, and C.~L. Zitnick, ``Microsoft coco: Common objects in
  context,'' in \emph{European conference on computer vision}.\hskip 1em plus
  0.5em minus 0.4em\relax Springer, 2014, pp. 740--755.

\bibitem{shapenet2015}
A.~X. Chang, T.~Funkhouser, L.~Guibas, P.~Hanrahan, Q.~Huang, Z.~Li,
  S.~Savarese, M.~Savva, S.~Song, H.~Su, J.~Xiao, L.~Yi, and F.~Yu,
  ``{ShapeNet: An Information-Rich 3D Model Repository}, Tech. Rep.
  arXiv:1512.03012 [cs.GR], 2015.

\bibitem{denninger_blenderproc_2019}
M.~Denninger, M.~Sundermeyer, D.~Winkelbauer, Y.~Zidan, D.~Olefir,
  M.~Elbadrawy, A.~Lodhi, and H.~Katam,
  ``\BIBforeignlanguage{en}{{BlenderProc}},''
  \emph{\BIBforeignlanguage{en}{arXiv:1911.01911 [cs]}}, 2019.

\bibitem{danielczuk_segmenting_2019}
M.~Danielczuk, M.~Matl, S.~Gupta, A.~Li, A.~Lee, J.~Mahler, and K.~Goldberg,
  ``\BIBforeignlanguage{en}{Segmenting {Unknown} {3D} {Objects} from {Real}
  {Depth} {Images} using {Mask} {R}-{CNN} {Trained} on {Synthetic} {Data}},''
  in \emph{\BIBforeignlanguage{en}{{IEEE} {International} {Conference} of
  {Robotics} and {Automation} ({ICRA})}}, 2019.

\bibitem{cheng2020hierarchical}
X.~Cheng, Y.~Zhong, M.~Harandi, Y.~Dai, X.~Chang, T.~Drummond, H.~Li, and
  Z.~Ge, ``Hierarchical neural architecture search for deep stereo matching,''
  \emph{arXiv preprint arXiv:2010.13501}, 2020.

\bibitem{bleyer_object_2011}
M.~Bleyer, C.~Rother, P.~Kohli, D.~Scharstein, and S.~Sinha,
  ``\BIBforeignlanguage{en}{Object stereo - {Joint} stereo matching and object
  segmentation},'' in \emph{\BIBforeignlanguage{en}{{IEEE} {Conference} on
  {Computer} {Vision} and {Pattern} {Recognition} ({CVPR})}}, 2011.

\bibitem{ladicky_joint_2010}
{\v L}.~Ladick{\'y}, P.~Sturgess, C.~Russell, S.~Sengupta, Y.~Bastanlar,
  W.~Clocksin, and P.~Torr, ``\BIBforeignlanguage{en}{Joint {Optimisation} for
  {Object} {Class} {Segmentation} and {Dense} {Stereo} {Reconstruction}},'' in
  \emph{\BIBforeignlanguage{en}{Procedings of the {British} {Machine} {Vision}
  {Conference}}}, 2010.

\bibitem{ferrari_segstereo_2018}
G.~Yang, H.~Zhao, J.~Shi, Z.~Deng, and J.~Jia,
  ``\BIBforeignlanguage{en}{{SegStereo}: {Exploiting} {Semantic} {Information}
  for {Disparity} {Estimation}},'' in \emph{\BIBforeignlanguage{en}{Computer
  {Vision} {\textendash} {ECCV}}}, 2018.

\bibitem{zhang_dispsegnet_2019}
J.~Zhang, K.~A. Skinner, R.~Vasudevan, and M.~Johnson-Roberson, ``{DispSegNet}:
  {Leveraging} {Semantics} for {End}-to-{End} {Learning} of {Disparity}
  {Estimation} {From} {Stereo} {Imagery},'' \emph{IEEE Robotics and Automation
  Letters}, 2019.

\bibitem{dovesi_real-time_2020}
P.~L. Dovesi, M.~Poggi, L.~Andraghetti, M.~Mart{\'i}, H.~Kjellstr{\"o}m,
  A.~Pieropan, and S.~Mattoccia, ``Real-{Time} {Semantic} {Stereo}
  {Matching},'' in \emph{{IEEE} {International} {Conference} on {Robotics} and
  {Automation}}, 2020.

\bibitem{he2016deep}
K.~He, X.~Zhang, S.~Ren, and J.~Sun, ``Deep residual learning for image
  recognition,'' in \emph{Proceedings of the IEEE conference on computer vision
  and pattern recognition}, 2016, pp. 770--778.

\bibitem{dosovitskiy2015flownet}
A.~Dosovitskiy, P.~Fischer, E.~Ilg, P.~Hausser, C.~Hazirbas, V.~Golkov, P.~Van
  Der~Smagt, D.~Cremers, and T.~Brox, ``Flownet: Learning optical flow with
  convolutional networks,'' in \emph{Proceedings of the IEEE international
  conference on computer vision}, 2015, pp. 2758--2766.

\bibitem{chen2017rethinking}
L.-C. Chen, G.~Papandreou, F.~Schroff, and H.~Adam, ``Rethinking atrous
  convolution for semantic image segmentation,'' \emph{arXiv:1706.05587 [cs]},
  2017.

\bibitem{wang2020axial}
H.~Wang, Y.~Zhu, B.~Green, H.~Adam, A.~Yuille, and L.-C. Chen, ``Axial-deeplab:
  Stand-alone axial-attention for panoptic segmentation,'' in \emph{European
  Conference on Computer Vision (ECCV)}, 2020.

\bibitem{sudre_dice_2017}
C.~H. Sudre, W.~Li, T.~Vercauteren, S.~Ourselin, and M.~Jorge~Cardoso,
  ``Generalised dice overlap as a deep learning loss function for highly
  unbalanced segmentations,'' in \emph{Deep Learning in Medical Image Analysis
  and Multimodal Learning for Clinical Decision Support}, 2017.

\bibitem{Kuhn1955thehungarian}
H.~W. Kuhn and B.~Yaw, ``The hungarian method for the assignment problem,''
  \emph{Naval Res. Logist. Quart}, 1955.

\bibitem{wong_3d_2018}
K.~C.~L. Wong, M.~Moradi, H.~Tang, and T.~Syeda-Mahmood,
  ``\BIBforeignlanguage{en}{{3D} {Segmentation} with {Exponential}
  {Logarithmic} {Loss} for {Highly} {Unbalanced} {Object} {Sizes}},'' in
  \emph{\BIBforeignlanguage{en}{Medical {Image} {Computing} and {Computer}
  {Assisted} {Intervention}}}, 2018.

\bibitem{huber1964robust}
P.~J. Huber, ``{Robust Estimation of a Location Parameter},'' \emph{The Annals
  of Mathematical Statistics}, 1964.

\bibitem{suchi2019easylabel}
M.~Suchi, T.~Patten, D.~Fischinger, and M.~Vincze, ``Easylabel: a
  semi-automatic pixel-wise object annotation tool for creating robotic rgb-d
  datasets,'' in \emph{International Conference on Robotics and Automation
  (ICRA)}, 2019.

\bibitem{rennie2016dataset}
C.~Rennie, R.~Shome, K.~E. Bekris, and A.~F. De~Souza, ``A dataset for improved
  rgbd-based object detection and pose estimation for warehouse
  pick-and-place,'' \emph{IEEE Robotics and Automation Letters}, 2016.

\bibitem{sui2018never}
Z.~Sui, Z.~Ye, and O.~C. Jenkins, ``\BIBforeignlanguage{en}{Never mind the
  bounding boxes, here's the sand filters},''
  \emph{\BIBforeignlanguage{en}{arXiv:1808.04969 [cs]}}, 2018.

\bibitem{ecins2016cluttered}
A.~{Ecins}, C.~{Fermüller}, and Y.~{Aloimonos}, ``Cluttered scene segmentation
  using the symmetry constraint,'' in \emph{IEEE International Conference on
  Robotics and Automation (ICRA)}, 2016.

\bibitem{rcvisard}
\BIBentryALTinterwordspacing
{Roboception GmbH}. [Online]. Available:
  \url{https://roboception.com/product/rc_visard-65-color/}
\BIBentrySTDinterwordspacing

\bibitem{zedstereo}
\BIBentryALTinterwordspacing
{Stereolabs Inc.} [Online]. Available: \url{https://www.stereolabs.com/zed/}
\BIBentrySTDinterwordspacing

\bibitem{calli2017yale}
B.~Calli, A.~Singh, J.~Bruce, A.~Walsman, K.~Konolige, S.~Srinivasa, P.~Abbeel,
  and A.~M. Dollar, ``Yale-cmu-berkeley dataset for robotic manipulation
  research,'' \emph{The International Journal of Robotics Research}, 2017.

\bibitem{song2016ssc}
S.~Song, F.~Yu, A.~Zeng, A.~X. Chang, M.~Savva, and T.~Funkhouser, ``Semantic
  scene completion from a single depth image,'' \emph{IEEE/CVF Conference on
  Computer Vision and Pattern Recognition}, 2017.

\bibitem{loshchilov2019decoupled}
I.~Loshchilov and F.~Hutter, ``\BIBforeignlanguage{en}{Decoupled {Weight}
  {Decay} {Regularization}},'' \emph{\BIBforeignlanguage{en}{arXiv:1711.05101
  [cs, math]}}, 2019.

\bibitem{xie2020uois3d}
C.~Xie, Y.~Xiang, A.~Mousavian, and D.~Fox, ``Unseen object instance
  segmentation for robotic environments,'' in \emph{arXiv:2007.08073}, 2020.

\bibitem{hodan2020bop}
T.~Hoda{\v{n}}, M.~Sundermeyer, B.~Drost, Y.~Labb{\'e}, E.~Brachmann,
  F.~Michel, C.~Rother, and J.~Matas, ``Bop challenge 2020 on 6d object
  localization,'' in \emph{Computer Vision -- ECCV 2020 Workshops}, 2020.

\bibitem{xu2020aanet}
H.~Xu and J.~Zhang, ``Aanet: Adaptive aggregation network for efficient stereo
  matching,'' in \emph{Proceedings of the IEEE/CVF Conference on Computer
  Vision and Pattern Recognition}, 2020.

\bibitem{MIFDB16}
N.~Mayer, E.~Ilg, P.~H{\"a}usser, P.~Fischer, D.~Cremers, A.~Dosovitskiy, and
  T.~Brox, ``A large dataset to train convolutional networks for disparity,
  optical flow, and scene flow estimation,'' in \emph{IEEE International
  Conference on Computer Vision and Pattern Recognition (CVPR)}, 2016.

\end{thebibliography}
